\documentclass[graybox]{svmult}

\usepackage{mathptmx}       
\usepackage{helvet}         
\usepackage{courier}        
\usepackage{type1cm}        
\usepackage{makeidx}         
\usepackage{graphicx}        
\usepackage{multicol}        
\usepackage[bottom]{footmisc}

\usepackage{amsmath}
\usepackage{graphicx}
\usepackage{mathrsfs}
\usepackage{amssymb}
\usepackage{verbatim}

\newcommand{\bx}{\mbox{\boldmath $x$}}
\newcommand{\bX}{\mbox{\boldmath $X$}}
\newcommand{\by}{\mbox{\boldmath $y$}}
\newcommand{\bY}{\mbox{\boldmath $Y$}}
\newcommand{\hR}{\hat{\mbox{$R$}}}

\newcommand{\hbtheta}{\hat{\mbox{\boldmath $\btheta$}}}

\newcommand{\hbbeta}{\hat{\mbox{\boldmath $\bbeta$}}}
\newcommand{\bmu}{\mbox{\boldmath $\mu$}}
\newcommand{\bSigma}{\mbox{\boldmath $\Sigma$}}

\newcommand{\btheta}{\mbox{\boldmath $\theta$}}
\newcommand{\bPsi}{\mbox{\boldmath $\Psi$}}
\newcommand{\bbeta}{\mbox{\boldmath $\beta$}}
\newcommand{\bxi}{\mbox{\boldmath $\xi$}}
\newcommand{\bkappa}{\mbox{\boldmath $\kappa$}}

\makeindex             

\begin{document}

\title*{Estimation of Classification Rules from Partially Classified Data}
\author{Geoffrey J.\ McLachlan and Daniel Ahfock}
\institute{G.J.\ McLachlan \at University of Queensland, Brisbane, 
\email{g.mclachlan@uq.edu.au}
\and D.\ Ahfock \at University of Queensland, Brisbane \email{d.ahfock@uq.edu.au}}
%
%
\maketitle

\abstract*{Each chapter should be preceded by an abstract (10--15 lines long) that summarizes the content. The abstract will appear \textit{online} at \url{www.SpringerLink.com} and be available with unrestricted access. This allows unregistered users to read the abstract as a teaser for the complete chapter. As a general rule the abstracts will not appear in the printed version of your book unless it is the style of your particular book or that of the series to which your book belongs.
Please use the 'starred' version of the new Springer \texttt{abstract} command for typesetting the text of the online abstracts (cf. source file of this chapter template \texttt{abstract}) and include them with the source files of your manuscript. Use the plain \texttt{abstract} command if the abstract is also to appear in the printed version of the book.}

\abstract{
We consider the situation where the observed sample contains 
some observations whose class of origin is known 
(that is, they are classified with respect to the 
$g$ underlying classes of interest), 
and where the remaining observations in the sample 
are unclassified (that is, their class labels are unknown). 
For class-conditional distributions taken to be known up
to a vector of unknown parameters, the aim is to estimate 
the Bayes' rule of allocation for the
allocation of subsequent unclassified observations. 
Estimation on the basis of both the classified and unclassified data
can be undertaken in a straightforward manner by fitting 
a $g$-component mixture model by maximum likelihood (ML)
via the EM algorithm in the situation where the observed data 
can be assumed to be an observed random sample 
from the adopted mixture distribution. 
This assumption applies if the missing-data mechanism is 
ignorable in the terminology pioneered by Rubin (1976). 
An initial likelihood approach was to use the 
so-called classification ML approach 
whereby the missing labels are taken to be parameters 
to be estimated along with the parameters of the 
class-conditional distributions. However, as it can lead to
inconsistent estimates, the focus of attention switched 
to the mixture ML approach 
after the appearance of the EM algorithm
(Dempster et al., 1977). 
Particular attention is given here to the
asymptotic relative efficiency (ARE) of the Bayes'
rule estimated from a partially classified sample.
Lastly, we consider briefly some recent results 
in situations where the missing label pattern
is non-ignorable for the purposes of ML estimation 
for the mixture model. 
}


\section{Introduction}
\label{sec:1}

We consider the estimation of a classifier from a sample that is not 
completely classified with respect to the predefined classes.
This problem goes back at least to the mid-seventies (McLachlan,
1975), and it received a boost shortly afterwards with
the advent of the EM algorithm (Dempster et al., 1977)
which could be applied to carry out maximum likelihood
(ML) estimation for a partially classified sample.
There is now a wide literature on the
formation of classifiers on the basis of a
partially classified sample
or semi-supervised learning (SSL) as it is
referred to in the machine learning literature.
In the sequel, it is assumed that the features with known
class labels are correctly classified, containing 
no misclassified features as, for example,
in McLachlan (1972) and, more recently,
Cannings and Samworth (2019).

More specifically, we focus on the case of $g=2$ classes $C_1$
and $C_2$ in which 
the $p$-dimensional feature vector $\bY$ measured on an entity 
is distributed as
\begin{equation}
\bY \sim N(\bmu_i,\bSigma)\quad {\rm in}\quad C_i\quad (i=1,2).
\label{eq:1}
\end{equation}
We let $\btheta$ contain the
$1+2p+{\textstyle\frac{1}{2}}p(p+1)$ unknown parameters,
consisting of the mixing proportion $\pi_1$, the
elements of the class means $\bmu_1$ and $\bmu_2$,
and the distinct elements of the common
class covariance matrix $\bSigma$.
The Bayes' rule of allocation $R(\by)$ in this case assigns
an entity with observed feature vector $\by$ to either
$C_1$ or $C_2$, according as 
$$d(\by)= \beta_0 +\bbeta^T \by$$
is greater or less than zero, where
\begin{eqnarray*}
\beta_0&=&-{\textstyle\frac{1}{2}}(\bmu_1+\bmu_2)^T
\bSigma^{-1}(\bmu_1-\bmu_2) + \log(\pi_1/\pi_2),\\
\bbeta&=&\bSigma^{-1}(\bmu_1-\bmu_2),
\end{eqnarray*}
and where $\pi_i$ denotes the prior probability
of membership of $C_i\,(i=1,2)$; 
see, for example, McLachlan (1992).

\section{History of SSL in Statistics}

In his discussion of the paper read to the Royal Statistical Society
by Hills (1966), Smith (1966) suggested that in the case of 
a completely unclassified sample 
which exhibits bimodality on some feature, 
a classifier be formed from the unclassified observations 
on the feature as follows: 
``One then arbitrarily divides them at the antimode, ....
On the basis of this division, we calculate a suitable allocation rule; 
and, by using this allocation rule, get an improved division, and so on. 
As far as I know, there is no theoretical research 
into the effect of `lifting oneself by one's own bootstraps' 
in this way.''

This led McLachlan (1975) to consider this approach 
as suggested by Smith (1966) 
under the normal homoscedastic model (1). 
Under the latter assumption, the procedure is equivalent to
treating the labels of the unclassified features as unknown parameters
to be estimated along with $\btheta$. 
This approach became subsequently known as the classification
maximum likelihood (CML) approach as  considered by Hartley and Rao (1968)
among others; see McLachlan and Basford (1988, Section 1.12).
The CML approach gives an inconsistent estimate of $\btheta$ except in special
cases like $\pi_1=\pi_2$.

In order to make the problem analytically tractable for the calculation
of the expected error rate of the estimated Bayes' rule,
McLachlan (1975) assumed that there were also
a limited number $n_{ic}$ of classified features available
from $C_i$ in
addition to the number of $n_u=n-n_c$ unclassified features, where
$n$ denotes the total size of the now partially classified sample
and $n_c=n_{1c}+n_{2c}$.

In the sequel, we let
$\bx_{\rm CC}=(\bx_1^T,\,\ldots,\,\bx_n^T)^T$
denote $n$ independent realizations of $\bX=(\bY^T, Z)^T$
as the completely classified training data,
where $Z$ denotes the
class membership of $\bY$, being equal to 1 if $\bY$
belongs to $C_1$, and zero otherwise.
We let $m_j$ be the missing-label indicator being equal to 1 if $z_j$
is missing and zero if it is available $(j=1,\,\ldots,\,n)$.
Accordingly, the unclassified sample $\bx_{\rm PC}$ is given by those
members $\bx_j$ in $\bx_{\rm CC}$ for which $m_j=0$ and only the feature
vectors $\by_j$ without their class labels $z_j$
for those members in $\bx_{\rm CC}$
for which $m_j=1$.

\section{Asymptotic expected error rate of CML approach}

In practice, $\bbeta$ has to be estimated from available training data.
It can be calculated iteratively as described in the previous section.
More formally, it can be obtained iteratively by
applying the expectation--maximization (EM)
algorithm of Dempster et al.\ (1977) with the following modification
(McLachlan,  1982).
Namely,
the E-step is executed using outright (hard) rather than
fractional (soft) assignment of each unclassified feature to a component 
of the mixture as with the standard application of the EM algorithm.
We let $\hbbeta_{\rm PC}^{(k)}$ denote the estimator after the $k$th iteration
of the vector $\bbeta=(\beta_0, \bbeta^T)^T$
of discriminant function coefficients obtained by the classification ML approach
applied to the partially classified sample $\bx_{\rm PC}$.
The estimated Bayes' rule using $\hbbeta_{\rm PC}^{(k)}$ for $\bbeta$ in the Bayes'
rule $R(\bbeta)$ is denoted by $\hR_{\rm PC}^{(k)}.$
The (overall) conditional error rate of 
$\hR_{\rm PC}^{(k)}$ is denoted by 
${\rm err}(\hbbeta_{\rm PC}^{(k)};\btheta)$.  

Then the expected  excess
error rate of the estimated Bayes' rule $\hR_{\rm PCC}^{(k)}$
is defined after the $k$th iteration by
$E\{{\rm err}(\hbbeta_{\rm PC}^{(k)};\btheta)\}-{\rm err}(\btheta),$
where ${\rm err}(\btheta)$ is the optimal error rate.

In the present SSL context, McLachlan (1975) showed in the case 
of equal, known prior probabilities 
that the overall expected error rate of this classifier
after the $k$th iteration
is given as, $n_u \rightarrow \infty$, by
\begin{equation}
E\{{\rm err}(\hbbeta_{\rm PC}^{(k)};\btheta)\}
=\Phi(-{\textstyle\frac{1}{2}}\Delta)+
\{\phi({\textstyle\frac{1}{2}}\Delta)/4\}\,a_1^{(k)} +O(n_c^{-2}),
\label{eq:2}
\end{equation}
where
\begin{eqnarray*}
a_1^{(k)}&=& h_1^{2k}\frac{\Delta}{4} +h_2^{2k}\frac{p-1}{\Delta}
(\frac{1}{n_{1c}}+\frac{1}{n_{2c}})+h_2^{2k}\frac{(p-1)\Delta}{n_c-2},\\
\\
h_1&=&\phi({\textstyle\frac{1}{2}})
[4\phi({\textstyle\frac{1}{2}})+\Delta\{1-2\Phi(-{\textstyle\frac{1}{2}})],\\
h_2&=&\{\phi({\textstyle\frac{1}{2}})\}^2\,(4+\Delta^2)/h_1,
\end{eqnarray*}
and where $\Delta=\{(\bmu_1-\bmu_2)^T\Sigma^{-1}(\bmu_1-\bmu_2)\}^{1/2}$ is the 
Mahalanobis distance between the class-conditional distributions and
${\rm err}(\btheta)=\Phi(-{\textstyle\frac{1}{2}}\Delta)$.

As it can be shown that both $|h_1|$ and $|h_2|$ are always less than one, it 
follows from (\ref{eq:2}) that the expected error rate of 
$\hR_{\rm PC}^{(k)}$ decreases after each iteration 
and converges to the optimal error rate as $k \rightarrow \infty$.

\section{Asymptotic relative efficiency of ML approach}

The construction of classifiers from partially classified data
can be undertaken also by the fitting of finite mixture models.
The ML estimate of the vector of parameters $\btheta$
can be obtained via the EM algorithm.
of Dempster et al.\ (1977).
As noted in McLachlan and Peel (2000),
it was the publication of this seminal paper that greatly 
stimulated interest in the use of finite mixture models.

We let
\begin{eqnarray}
\log L_{\rm C}(\btheta)&=& 
\sum_{j=1}^n(1-m_j)[z_j\log\{\pi_1 \phi(\by_j;\bmu_1,\bSigma)\}
+(1-z_j) \log\{\pi_2 \phi(\by_j;\bmu_2,\bSigma)\}]\nonumber\\
\label{eq:4}\\
\log L_{\rm UC}(\btheta)&=&  
\sum_{j=1}^n m_j \log \sum_{i=1}^2 \pi_i \phi(\by_j;\bmu_i, \Sigma),\label{eq:5}\\
\log L_{\rm PC}^{(\rm ig)}(\btheta)&=&\log L_{\rm C}(\btheta)
+\log L_{\rm UC}(\btheta),
\label{eq:5}
\end{eqnarray}
where $\phi(\by_j;\bmu,\bSigma)$ denotes the multivariate normal
density with mean $\bmu$ and covariance matrix $\bSigma$.
In situations where one proceeds by ignoring
the ``missingness'' of the class labels,
$L_{\rm C}(\btheta)$ and $L_{\rm UC}(\btheta)$
denote the likelihood function formed from the classified data
and the unclassified data, respectively, and
$L_{\rm PC}^{(\rm ig)}(\btheta)$ is the likelihood function
formed from the partially classified sample $\bx_{\rm PC}$.
The log of the likelihood $L_{\rm CC}(\btheta)$ for the
completely classified sample $\bx_{\rm CC}$ is given by (\ref{eq:4})
with all $m_j=0$.

Situations in the present context where it is appropriate to 
ignore the missing-data mechanism in carrying out likelihood inference
are where the
missing labels are missing at random in the framework for
missing data  pioneered by Rubin (1976).
This will be the case in the present context  
if the missingness of the labels does not depend on
the features nor the labels (missing completely at random)
or if the missingness depends only on the features (missing at random),
as in McLachlan and Gordon (1989).

We let $\hbtheta_{\rm CC}$ and $\hbtheta_{\rm PC}$
be the estimate of $\btheta$ formed 
by consideration of $L_{\rm CC}(\btheta)$ and $L_{\rm PC}^{(\rm ig)}(\btheta)$,
respectively, and we let
$\hbbeta_{\rm CC}$ and $\hbbeta_{\rm PC}^{(\rm ig)}$ be the estimates of
$\bbeta$ formed from the elements of
$\hbtheta_{\rm CC}$ and $\hbtheta_{\rm PC}^{(\rm ig)}$, respectively.
The relative efficiency of the estimated Bayes' rule
$R(\hbbeta_{\rm PC}^{(\rm ig)}$ compared to the rule $\hR_{\rm CC}$
using $\hbbeta_{\rm CC}$
for $\bbeta$ based on the completely classified sample $\bx_{\rm CC}$
is defined by
\begin{equation}
{\rm ARE}(R_{\rm PC}^{(\rm ig)})
=\frac{E\{{\rm err}(\hbbeta_{\rm CC};\btheta)\} -{\rm err}(\btheta)}
{E\{{\rm err}(\hbbeta_{\rm PC}^{(\rm ig)};\bbeta)\}-{\rm err}(\btheta)},
\label{eq:6}
\end{equation}
where the expectation in the numerator and denominator of
the right-hand side of (\ref{eq:6})
is taken over the distribution of the estimators of $\bbeta$
and is expanded up to terms of the first order.

Under the assumption that 
the class labels are 
missing always completely at random,
(that is, the missingness of the labels does not depend on the data),
Ganesalingam and McLachlan (1978) derived the 
ARE of $\hR_{\rm PC}^{(\rm ig)}$
compared to $\hR_{\rm CC}$
in the case of a completely unclassified sample $(\gamma = n_u/n= 1)$ 
for univariate features $(p = 1)$.
Their results are listed in Table~\ref{tab:1} 
for $\Delta = 1, 2, 3,$ and 4. 
O'Neill (1978) extended their result to
multivariate features and for arbitrary $\gamma$
using the result of Efron (1975) for the information matrix
of $\bbeta$  in applying logistic regression.
His results showed that this ARE was not sensitive to the
values of $p$ and does not vary with $p$ for equal class
prior probabilities. 
Not surprisingly, it can be seen from Table \ref{tab:1}
that the ARE of $\hR_{\rm PC}^{(\rm ig)}$
for a totally unclassified
sample is low, particularly for classes weakly separated as 
represented by $\Delta= 1$ in Table~1.

\begin{table}
\caption{Asymptotic relative efficiency of $\hR_{\rm PC}^{(\rm ig)}$ 
compared to $\hR_{\rm CC}$}
\label{tab:1}       
%
%
\begin{tabular}{p{1.5cm}p{2.2cm}p{2cm}p{2cm}p{2cm}}
\hline\noalign{\smallskip}
$\pi_1$ & $\Delta=1$ &  $\Delta=2$& $\Delta=3$ &  $\Delta=4$ \\
\noalign{\smallskip}\svhline\noalign{\smallskip}
0.1 & 0.0036 & 0.0591 & 0.2540  & 0.5585\\
0.2 & 0.0025 & 0.0668 & 0.2972  & 0.6068 \\
0.3 & 0.0027 & 0.0800 & 0.3289  & 0.6352 \\
0.4 & 0.0038 & 0.0941 & 0.3509  & 0.6522 \\
0.5 & 0.0051 & 0.1008 & 0.3592  & 0.6580 \\
\noalign{\smallskip}\hline\noalign{\smallskip}
\end{tabular}
\end{table}

In other work on the ARE of $\hR_{\rm PC}^{(\rm ig)}$
compared to $\hR_{\rm CC}$,
McLachlan and Scot  (1995)
evaluated it where the unclassified
univariate features had labels
missing always at random
due to truncation of the features.

\section{Modelling missingness for unobserved class labels}

In many practical applications class labels will be assigned 
by experts. 
Manual annotation of the dataset can induce 
a systematic missingness mechanism. 
This led Ahfock and McLachlan (2019a,b) to pursue the 
idea that the probability that a particular feature is unlabelled 
is related to the difficulty of determining its true class label. 
As an example, suppose medical professionals are asked to classify 
each image from a set of MRI scans into three groups, 
tumour present, no tumour present, or unknown.
It seems reasonable to expect that the unassigned images 
will correspond to those that do not present clear evidence 
for the presence or absence of a tumour. 
The unlabelled images will exist in regions of the feature space 
where there is class overlap.
In these situations, the unlabelled features carry additional 
information that can be used to improve the efficiency 
of parameter estimation. 

The missing-data mechanism of Rubin (1976) is specified
in the present context by the conditional distribution
\begin{equation}
{\rm pr}\{M_j=m_j\mid \by_j, \,z_j;\bkappa\} \quad (j=1,\,\ldots,\,n),
\label{eq:9}
\end{equation}
where $\bkappa$ is a vector of parameters.
Ahfock and McLachlan (2019a,b) proposed that 
\begin{eqnarray}
{\rm pr}\{M_j=1\mid \by_j,z_j\}&=& {\rm pr} \{M_j=1\mid \by_j\}\nonumber\\
&=&q(\by_j;\btheta,\bxi),
\label{eq:10}
\end{eqnarray}
where the parameter $\bxi$ is distinct from $\btheta$.
On putting $\bPsi=(\btheta^T,\bxi^T)^T$,
an obvious choice for the function $q(\by_j;\bPsi)$
is the logistic model
\begin{equation}
q(\by_j;\bPsi)= \frac{\exp\{\xi_0+\xi_1 e_j\}}
{1+\exp\{\xi_0+\xi_1 e_j\}},
\label{eq:11}
\end{equation}
where
\begin{equation}
e_j=-\sum_{i=1}^2 \tau_i(\by_j;\btheta)\log \tau_i(\by_j;\btheta)
\label{eq:12}
\end{equation}
denotes the entropy for $\by_j$, and where
$\tau_i(\by_j;\btheta)$
is the posterior probability that the $j$th entity with observed 
feature $\by_j$  belongs to Class $C_i\,(1=1,2)$.

The log of the full likelihood function for $\bPsi$
is given by
\begin{equation}
\log L_{\rm PC}^{(\rm full)}(\bPsi)=
\log L_{\rm PC}^{(\rm ig)}(\btheta) +
\log L_{\rm PC}^{(\rm miss)}(\bPsi),
\label{eq:13}
\end{equation}
where
\begin{eqnarray}
\log L_{\rm PC}^{(\rm miss)}(\bPsi)&=&
\sum_{j=1}^n [(1-m_j)\log \{1- q(\by_j;\bPsi)\}
+ \, m_j \log q(\by_j;\bPsi)]
\label{eq:15}
\end{eqnarray}
is the log likelihood function for $\bPsi$
formed on the basis of the missing-label indicators
$m_j\,(j=1,\,\ldots,\,n)$.

Under the model (\ref{eq:11}) for the missingness of the class labels
Ahfock and McLachlan (2019a,b)
showed for the normal homoscedastic model (\ref{eq:1})
that the ARE of the Bayes' rule using the full ML estimate of $\bbeta$
can not only be improved 
but, rather surprisingly, can be greater than one.

\section{Fractionally supervised classification}

In this section we make use of the model (\ref{eq:11}) to examine
the potential usefulness of fractionally supervised classification (FSC)
as proposed by Vrbik and McNicholas (2015) and considered further
by Gallaugher and McNicholas (2019). With this approach, the 
patameter $\btheta$ is estimated by consideration of the objective function
$L_{\rm PC}^{(\alpha)}(\btheta)$ defined for a given $\alpha$ in [0,1]  by
$$\log L_{\rm PC}^{(\alpha)}(\btheta)=\alpha \log L_{\rm C}(\btheta) +
(1-\alpha) \log L_{\rm UC}(\btheta).$$
One suggestion for the choice of $\alpha$ in practice is to use BIC
(Schwarz, 1978).

\begin{figure}
\sidecaption
\includegraphics[scale=.25]{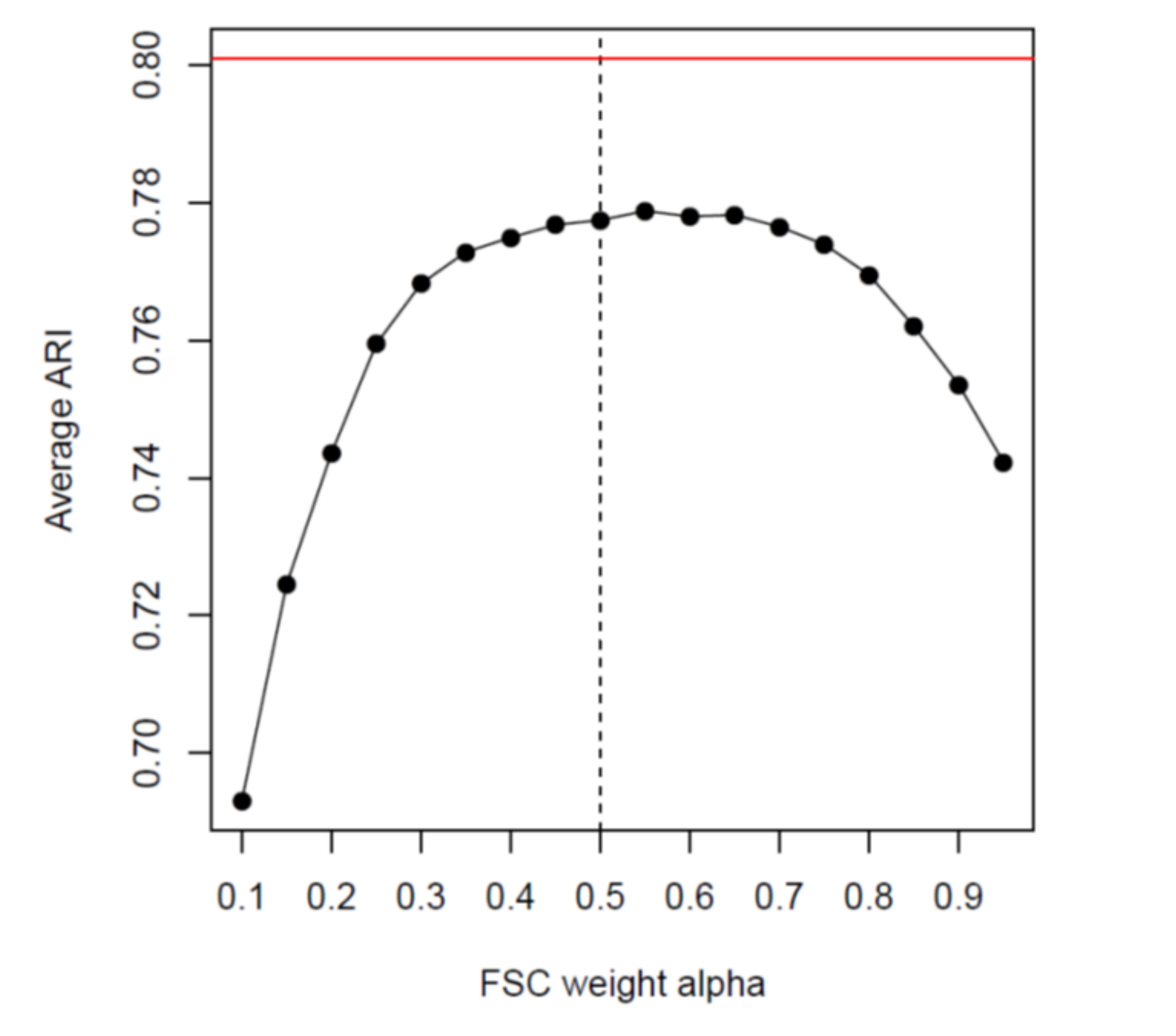}
\caption{Plot of simulated ARI for 
various values of $\alpha$}
\label{fig:1} 
\end{figure}

We report here a 
simulation experiment undertaken by Ahfock and McLachlan (2019a)
in which a partially classified sample of size $n=500$
was generated on each of $N=100$ replications.
Bivariate features were generated  from a mixture of two normal bivariate
distributions in equal proportions $(\pi_1=\pi_2=0.5)$ with unequal
covariance matrices,
where the two components correspond to $g=2$ classes.
The component means were given by $\bmu_1=(0,0)^T$ and $\bmu_2=(0,3)^T$
with the component-covariance matrices having unit variances for both variables
with correlation 0.7 in the first component and zero correlation
in the second component.
The conditional distribution of the missing-label indicators $M_j$
was specified by the model (\ref{eq:11}) with 
$\xi_0=-5$ and $\xi_1=100$.
For each partially classified sample $\bx_{\rm PC}$ generated,
the estimate $\hbtheta_{\rm PC}^{(\alpha)}$ of $\btheta$ 
was calculated via maximization of the objective function
$L_{\rm PC}^{(\alpha)}$ for a grid of values of $\alpha$,
along with the estimate $\hbtheta_{\rm PC}^{(\rm full)}$
using the full likelihood function $L_{\rm PC}^{(\rm full)}(\bPsi)$.
On each replication the adjusted Rand index (ARI) 
for the estimated Bayes' rule was obtained 
by applying it to 2,000 data points in a test set.
The average values of these ARI's are displayed in Figure~\ref{fig:1}.
They show that as $\alpha$ moves away from a small 
neighbourhood of $\alpha=0.5$, the performance of the rule using the
fractionally supervised estimate falls dramatically.
The horizontal line in Figure~\ref{fig:1} 
is the simulated value of the ARI for the use
of $\hbtheta_{\rm PC}^{(\rm full)}$.

\begin{thebibliography}{99.}%
\bibitem{science-journal} 
Ahfock, D., McLachlan, G.J.:
On missing data patterns in semi-supervised learning. 
ePreprint arXiv:1904.02883 (2019a) 
\bibitem{science-journal}
Ahfock, D., McLachlan, G.J.:
An apparent paradox: A classifier trained 
from a partially classified sample may have smaller 
expected error rate than that if the sample were 
completely classified. 
ePreprint arXiv:1910.09189v2 (2019b)
\bibitem{science-journal}
Cannings, T.I., Fan, Y., Samworth, R.J.:
Classification with imperfect training labels.
Biometrika \textbf{107},. To appear (2020)
\bibitem{science-journal} 
Dempster, A.P., Laird, N.M., Rubin, D.B.: 
Maximum likelihood from incomplete data via the EM algorithm
(with discussion). 
J. R. Statist. Soc. B \textbf{39}, 1--22 (1977)
\bibitem{science-journal} 
Efron, B.: 
The efficiency of logistic regression compared
to normal discriminant analysis. 
J.  Amer. Statist. Assoc. \textbf{70}, 892--898 (1975)
\bibitem{science-journal} 
Gallaugher, M., McNicholas, P.D.:
On fractionally-supervised classification: 
weight selection and extension to the multivariate $t$--distribution. 
J. Classification \textbf{36}, 232--265,  (2019)
\bibitem{science-journal} 
Ganesalingam, S., McLachlan, G.J.
The efficiency of a linear discriminant function based 
on unclassified initial samples. 
Biometrika \textbf{65}, 658--665 (1978)
\bibitem{science-journal} 
Hartley, H.O. Rao, J.N.K.:
Classification and estimation in analysis of variance problems.
Int. Statist. Rev. \textbf{36}, 141--147 (1968)
\bibitem{science-journal} 
Hills, M.: Allocation rules and their error rates (with discussion).
J. R. Statist. Soc. B \textbf{28}, 1--31 (1966)
\bibitem{science-journal} 
McLachlan, G.J.  
Asymptotic results for discriminant analysis 
when the initial samples are misclassified. 
Technometrics \textbf{14}, 415--422 (1972) 
\bibitem{science-journal} 
McLachlan, G.J.: 
The classification and mixture maximum likelihood 
approaches to cluster analysis. 
In: Krishnaiah, P.A. Kanal, L. (eds.)
Handbook of Statistics Vol.\ 2,
North-Holland, pp.\ 199--208. Amsterdam (1982)
\bibitem{science-journal} 
McLachlan, G.J.: 
Discriminant Analysis and Statistical Pattern Recognition. 
Wiley, New York (1992)
\bibitem{science-journal} 
McLachlan, G.J.: 
Iterative reclassification procedure for
constructing and asymptotically optimal rule 
of allocation in discriminant analysis. 
J. Amer. Statist. Assoc. \textbf{70}, 365--369 (1975)
\bibitem{science-journal} 
McLachlan, G.J., Basford, K.E.:
Mixture Models: Inference and Applications to Clustering.
Marcel Dekker, New York (1988)
\bibitem{science-journal} 
McLachlan, G.J., Gordon, R.D.:
Mixture models for partially unclassified data: 
a case study of renal venous renin levels in essential hypertension. 
Statist. Med. \textbf{8}, 1291--1300 (1989)
\bibitem{science-journal} 
McLachlan, G.J., Peel, D.:
Finite Mixture Models.
Wiley, New York, (2000)
\bibitem{science-journal} 
McLachlan, G.J., Scot, D.:
On the asymptotic relative efficiency of the linear discriminant function 
under partial nonrandom classification of the training data. 
Statist. Comp. Simul. \textbf{52}, 452--456 (1995)
\bibitem{science-journal} 
O'Neill, T.J.: 
Normal discrimination with unclassified observations. 
J. Amer. Statist. Assoc. \textbf{73}, 821--826 (1978)
\bibitem{science-journal} 
Rubin, D.B.: 
Inference and missing data. Biometrika \textbf{63}, 581--592 (1976)
\bibitem{science-journal} 
Schwarz, G.:
Estimating the dimension of a model.
Ann. Statist. \textbf{6}, 461--464 (1978)
\bibitem{science-journal} 
Smith, C.A.B.:
Contribution to the discussion of paper by M.\ Hills.
J. R. Statist. Soc. \textbf{28}, 21 (1966)
\bibitem{science-journal} 
Vrbik, I., McNicholas, P. D.:
Fractionally-supervised classification. 
J. Classification \textbf{32}, 359--381 (2015)
%
\end{thebibliography}
\end{document}